# Image Segmentation Algorithms Overview

*Song Yuheng[1], Yan Hao[1]*

(1. SiChuan University, SiChuan, ChengDu)

**Abstract：** The technology of image segmentation is widely used in medical image processing, face recognition pedestrian detection, etc. The current image segmentation techniques include region-based segmentation, edge detection segmentation, segmentation based on clustering, segmentation based on weakly-supervised learning in CNN, etc. This paper analyzes and summarizes these algorithms of image segmentation, and compares the advantages and disadvantages of different algorithms. Finally, we make a prediction of the development trend of image segmentation with the combination of these algorithms.

## 1 Introduction

An image is a way of transferring information, and the image contains lots of useful information. Understanding the image and extracting information from the image to accomplish some works is an important area of application in digital image technology, and the first step in understanding the image is the image segmentation. In practice, it is often not interested in all parts of the image, but only for some certain areas which have the same characteristics[1]. Image segmentation is one of the hotspots in image processing and computer vision. It is also an important basis for image recognition. It is based on certain criteria to divide an input image into a number of the same nature of the category in order to extract the area which people are interested in. And it is the basis for image analysis and understanding of image feature extraction and recognition.

There are many commonly used image segmentation algorithms. This paper mainly describes the following five algorithms for simple analysis. The first is the threshold segmentation method. Threshold segmentation is one of the most commonly used segmentation techniques in region-based segmentation algorithms[2]. Its essence is to automatically determine the optimal threshold according to a certain criterion, and use these pixels according to the gray level to achieve clustering. Followed by the regional growth segmentation. The basic idea of the regional growth algorithm is to combine the pixels with similar properties to form the region, that is, for each region to be divided first to find a seed pixel as a growth point, and then merge the surrounding neighborhood with similar properties of the pixel in its area. Then is the edge detection segmentation method. Edge detection segmentation algorithm refers to the use of different regions of the pixel gray or color discontinuity detection area of the edge in order to achieve image segmentation[3]. The next is the segmentation based on clustering. The algorithm based on clustering is based on the similarity between things as the criterion of class division, that is, it is divided into several subclasses according to the internal structure of the sample set, so that the same kind of samples are as similar as possible, and the different are not as similar as possible[4]. The last is the segmentation based on weakly-supervised learning in CNN. It refers to the problem of assigning a semantic label to every pixel in the image and consists of three parts. 1)Give an image which contains which objects. 2)Give the border of an object. 3) The object area in the image is marked with a partial pixel[5].

At present, from the international image segmentation method, the specific operation of the process of segmentation method is very diverse and complex, and there is no recognized a unified standard. This paper discusses and compares the above four methods, and learns from the shortcomings to analyze better solutions and make future forecasts.

## 2 Analysis

### 2.1 Region-based Segmentation

#### 2.1.1 Threshold Segmentation

Threshold segmentation is the simplest method of image segmentation and also one of the most common parallel segmentation methods. It is a common segmentation algorithm which directly divides the image gray scale information processing based on the gray value of different targets. Threshold segmentation can be divided into local threshold method and global threshold method. The global threshold method divides the image into two regions of the target and the background by a single threshold[6]. The local threshold method needs to select multiple segmentation thresholds and divides the image into multiple target regions and backgrounds by multiple thresholds.

The most commonly used threshold segmentation algorithm is the largest interclass variance method (Otsu)[7], which selects a globally optimal threshold by maximizing the variance between classes. In addition to this, there are entropy-based threshold segmentation method, minimum error method, co-occurrence matrix method, moment preserving method, simple statistical method, probability relaxation method, fuzzy set method and threshold methods combined with other methods[8].

The advantage of the threshold method is that the calculation is simple and the operation speed is faster. In particular, when the target and the background have high contrast, the segmentation effect can be obtained. The disadvantage is that it is difficult to obtain accurate results for image segmentation problems where there is no significant gray scale difference or a large overlap of the gray scale values in the image[9]. Since it only takes into account the gray information of the image without considering the spatial information of the image, it is sensitive to noise and grayscale unevenness, leading it often combined with other methods[10].

### 2.1.2 Regional Growth Segmentation

The regional growth method is a typical serial region segmentation algorithm, and its basic idea is to have similar properties of the pixels together to form a region[11]. The method requires first selecting a seed pixel, and then merging the similar pixels around the seed pixel into the region where the seed pixel is located.

Figure 1 shows an example of a known seed point for region growing. Figure 1 (a) shows the need to split the image. There are known two seed pixels (marked as gray squares) which are prepared for regional growth. The criterion used here is that if the absolute value of the gray value difference between the pixel and the seed pixel is considered to be less than a certain threshold T, the pixel is included in the region where the seed pixel is located. Figure 1 (b) shows the regional growth results at T = 3, and the whole plot is well divided into two regions. Figure 1 (c) shows the results of the region growth at T = 6 and the whole plot is in an area. Thus the choice of threshold is very important[12].

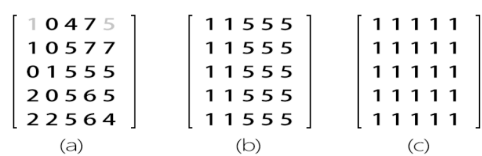

Fig. 1：Examples of regional growth

The advantage of regional growth is that it usually separates the connected regions with the same characteristics and provides good boundary information and segmentation results. The idea of regional growth is simple and requires only a few seed points to complete. And the growth criteria in the growing process can be freely specified. Finally, it can pick multiple criteria at the same time. The disadvantage is that the computational cost is large[13]. Also the noise and grayscale unevenness can lead to voids and over-division. The last is the shadow effect on the image is often not very good[14].

### 2.2 Edge Detection Segmentation

The edge of the object is in the form of discontinuous local features of the image, that is, the most significant part of the image changes in local brightness, such as gray value of the mutation, color mutation, texture changes and so on[15]. The use of discontinuities to detect the edge, so as to achieve the purpose of image segmentation.

There is always a gray edge between two adjacent regions with different gray values in the image, and there is a case where the gray value is not continuous. This discontinuity can often be detected using derivative operations, and derivatives can be calculated using differential operators[16]. Parallel edge detection is often done by means of a spatial domain differential operator to perform image segmentation by convoluting its template and image. Parallel edge detection is generally used as a method of image preprocessing. The widely first-order differential operators are Prewitt operator, Roberts operator and Sobel operator[17]. The second-order differential operator has nonlinear operators such as Laplacian, Kirsch operator and Wallis operator.

### 2.2.1 Sobel Operator

The Sobel operator is mainly used for edge detection, and it is technically a discrete differential operator used to calculate the approximation of the gradient of the image luminance function. The Sobel operator is a typical edge detection operator based on the first derivative. As a result of the operator in the introduction of a similar local average operation, so the noise has a smooth effect, and can effectively eliminate the impact of noise. The influence of the Sobel operator on the position of the pixel is weighted, which is better than the Prewitt operator and the Roberts operator.

The Sobel operator consists of two sets of 3x3 matrices, which are transverse and longitudinal templates, and are plotted with the image plane, respectively, to obtain the difference between the horizontal and the longitudinal difference. In actual use, the following two templates are used to detect the edges of the image.

$$Gx = \begin{bmatrix} -1 & 0 & 1 \\ -2 & 0 & 2 \\ -1 & 0 & 1 \end{bmatrix}$$

Detect horizontal edge (transverse template)

$$Gy = \begin{bmatrix} 1 & 2 & 1 \\ 0 & 0 & 0 \\ -1 & -2 & -1 \end{bmatrix}$$

Detect vertical edge (longitudinal template)

The horizontal and vertical gradient approximations of each pixel of the image can be combined to calculate the size of the gradient using the following formula:

$$G = \sqrt[2]{G_x^2 + G_y^2}$$

The gradient can then be calculated using the following formula:

$$\Theta = \arctan(\frac{G_y}{G_x})$$

In the above example, if the above angle Θ is equal to zero, that is, the image has a longitudinal edge, and the left is darker than the right.

#### 2.2.2 Laplacian Operator

Laplace operator is an isotropic operator, second order differential operator. It is more appropriate when it is only concerned with the position of the edge regardless of the pixel gray scale difference around it[18]. The Laplace operator's response to isolated pixels is stronger than the edge or line, and therefore applies only to noise-free images. In the presence of noise, the Laplacian operator needs to perform low-pass filtering before detecting the edge. Therefore, the usual segmentation algorithm combines the Laplacian operator with the smoothing operator to generate a new template.

Laplacian operator is also the simplest isotropic differential operator with rotational invariance. The Laplace transform of a two-dimensional image function is an isotropic second derivative, which is more suitable for digital image processing, and the pull operator is expressed as a discrete form:

$$\nabla^2 f = \frac{\partial^2 f}{\partial x^2} + \frac{\partial^2 f}{\partial y^2}$$

In addition, the Laplace operator can also be expressed in the form of a template.

$$\begin{bmatrix} 0 & 1 & 0 \\ 1 & -4 & 1 \\ 0 & 1 & 0 \end{bmatrix}$$

Discrete Laplacian garlic template

$$\begin{bmatrix} 1 & 1 & 1 \\ 1 & -8 & 1 \\ 1 & 1 & 1 \end{bmatrix}$$

Extended template

The Laplacian operator is used to improve the blurring effect due to the blurring effect, since it conforms to the descent model. Diffusion effect is often occurring in the imaging process.

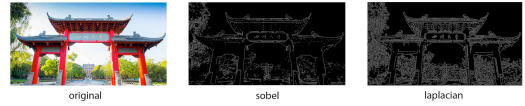

Fig. 2: Examples of different operator image show

### 2.3 Segmentation based on clustering

There is no general theory of image segmentation. However, with the introduction of many new theories and methods of various disciplines, there have been many image segmentation methods combined with some specific theories and methods. The so-called class, refers to the collection of similar elements. Clustering is in accordance with certain requirements and laws of the classification of things in the process[19]. The feature space clustering method is used to segment the pixels in the image space with the corresponding feature space points. According to their aggregation in the feature space, the feature space is segmented, and then they are mapped back to the original image space to get the segmentation result.

K-means is one of the most commonly used clustering algorithm. The basic idea of K-means is to gather the samples into different clusters according to the distance. The closer the two points are, the closer they are to get the compact and independent clusters as clustering targets[20]. The implementation process of K-means is expressed as follows:

(1) Randomly select K initial clustering centers;

(2) Calculate the distance from each sample to each cluster center, and return each sample to the nearest clustering center;

(3) For each cluster, with the mean of all samples as the cluster of new clustering centers;

(4) Repeat steps (2) to (3) until the cluster center no longer changes or reaches the set number of iterations[21].

The advantage of K-Means clustering algorithm is that the algorithm is fast and simple, and it is highly efficient and scalable for large data sets. And its time complexity is close to linear, and suitable for mining large-scale data sets. The disadvantage of K-means is that its clustering number K has no explicit selection criteria and is difficult to estimate[22]. Secondly, it can be seen from the K-means algorithm framework that every iteration of the algorithm traverses all the samples, so the time of the algorithm is very expensive. Finally, the K-means algorithm is a distance-based partitioning method[23]. It is only applicable to the data set which is convex and not suitable for clustering non-convex clusters.

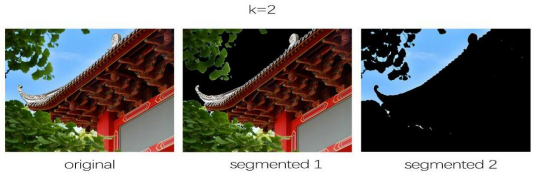

Fig. 3: Examples of different K-means image show (k = 2)

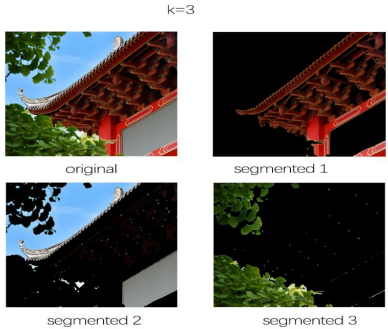

Fig. 4: Examples of different K-means image show (k = 3)

## 2.4 Segmentation based on weakly-supervised learning in CNN

In recent years, the deep learning has been in the image classification, detection, segmentation, high-resolution image generation and many other areas have made breakthrough results[24]. In the aspect of image segmentation, an algorithm is proposed which is more effective in this field, which is the weakly- and semi-supervised learning of a DCNN for semantic image segmentation. Google's George Papandreou and UCLA's Liang-Chieh Chen studied the use of bounding box and image-level labels as markup training data on the basis of DeepLab and used the expected maximization algorithm (EM) to estimate unmarked Pixel class and CNN parameters. DeepLab method is divided into two steps, the first step is still using the FCN to get the coarse score map and interpolate to the original image size, and then the second step began to borrow the fully connected CRF from the FCN to get the details of the segmentation refinement[25].

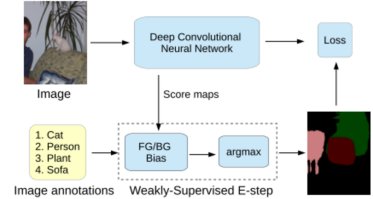

Fig. 5: DeepLab model training using image-level labels

For the image-level tagged data, we can observe the pixel value x of the image and the mark z of the image level, but do not know the label y for each pixel, so y is treated as a hidden variable. Use the following probability graph mode:

$$P(\boldsymbol{x},\boldsymbol{y},\boldsymbol{z};\boldsymbol{\theta}) = P(\boldsymbol{x})\left(\prod_{m=1}^{M} P(y_m|\boldsymbol{x};\boldsymbol{\theta})\right) P(\boldsymbol{z}|\boldsymbol{y}).$$

Use the EM algorithm to estimate $\theta$ and y. E step is fixed $\theta$ y expect the value of y, and M step is fixed y using SGD to calculate $\theta$[26].

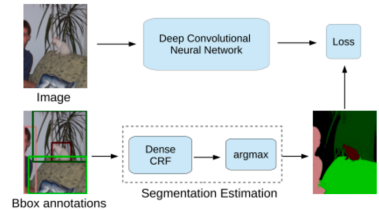

Fig. 5: DeepLab model training from bounding boxes

For the training image that gives the bounding box mark, the method uses the CRF to automatically segment the training image, and then do full supervision on the basis of the segmentation. Experiments show that simply using the image level of the mark to get the segmentation effect is poor, but the use of bounding box training data can get better results[27].

## 3 Conclusion

As can be seen from the paper, it is found that it is difficult to find a segmentation way to adapt with all images. At present, the research of image segmentation theory is not perfect, and there are still many practical problems in applied research. Through comparing the advantages and disadvantages of the various image segmentation algorithms, the development of image segmentation techniques may present the following trends: 1) The combination of multiple segmentation methods. Because of the diversity and uncertainty of the image, it is necessary to combine the multiple segmentation methods and make full use of the advantages of different algorithms on the basis of multi-feature fusion, so as to achieve better segmentation effect. 2) In the parameter selection using machine learning algorithm for analysis, in order to improve the segmentation effect. Such as the threshold selection in threshold segmentation and the selection of K values in the K-means algorithm. 3) CNN model is used to frame the ROI, and then segmented by non-machine learning segmentation method to improve the segmentation effect. It is believed that in the future research and exploration, there will be more image segmentation method to be further developed and more widely used.